\documentclass[11pt,a4paper]{article}
\usepackage[T1]{fontenc}
\usepackage[utf8]{inputenc}
\usepackage{lmodern}
\usepackage[a4paper,top=0.96in,bottom=0.92in,left=0.99in,right=0.99in,headheight=14pt]{geometry}
\usepackage{microtype}
\usepackage{amsmath,amssymb}
\usepackage{booktabs,tabularx,array}
\usepackage{enumitem}
\usepackage{fancyhdr}
\usepackage{titlesec}
\usepackage{xcolor}
\usepackage{hyperref}
\usepackage{xurl}
\usepackage{listings}
\usepackage{caption}
\usepackage{float}
\usepackage{graphicx}
\usepackage{tikz}
\usetikzlibrary{arrows.meta,positioning}

\definecolor{navy}{HTML}{1E4063}
\definecolor{muted}{HTML}{586474}
\definecolor{light}{HTML}{EDF3F7}
\hypersetup{colorlinks=true,linkcolor=navy,citecolor=navy,urlcolor=navy,
  pdftitle={A Human-Verified AI Workflow for the Kenya Forest Sounds Dataset (KFSD)},
  pdfauthor={Barasa Michael Murunga; Allan Vikiru}}
\setlist[itemize]{leftmargin=1.38em,itemsep=2.2pt,parsep=0pt,topsep=3pt}
\setlist[enumerate]{leftmargin=1.55em,itemsep=2.2pt,parsep=0pt,topsep=3pt}
\titleformat{\section}{\large\bfseries\color{navy}}{\thesection}{0.67em}{}
\titleformat{\subsection}{\normalsize\bfseries}{\thesubsection}{0.63em}{}
\titleformat{\subsubsection}{\normalsize\itshape}{\thesubsubsection}{0.62em}{}
\titlespacing*{\section}{0pt}{16pt plus 2pt}{6pt}
\titlespacing*{\subsection}{0pt}{11pt plus 2pt}{4pt}
\titlespacing*{\subsubsection}{0pt}{8pt plus 1pt}{3pt}
\lstdefinestyle{paper}{
 basicstyle=\ttfamily\footnotesize,breaklines=true,breakatwhitespace=false,
 xleftmargin=9pt,xrightmargin=7pt,frame=single,framerule=0.35pt,
 rulecolor=\color{gray!35},backgroundcolor=\color{gray!4},
 showstringspaces=false,columns=fullflexible,keepspaces=true,
 aboveskip=8pt,belowskip=7pt,tabsize=4
}
\begin{document}
\begin{center}
{\parbox{0.96\linewidth}{\centering\fontsize{17}{21}\selectfont\bfseries A Human-Verified AI Workflow for Annotating Environmental Sound Data}\par}
\vspace{13pt}
{\normalsize Barasa Michael Murunga$^{1}$ \quad Allan Vikiru$^{1,*}$\par}
\vspace{5pt}
{\small $^{1}$ Conservation Technology Research Lab, Strathmore University, Ole Sangale Road, Nairobi, Kenya\par}
{\footnotesize Emails: \texttt{michael.barasa@strathmore.edu}; \texttt{avikiru@strathmore.edu}$^{*}$\par}
{\footnotesize $^{*}$ Corresponding author\par}
\end{center}
\vspace{3pt}
\begin{abstract}
Large-scale annotation of environmental sound recordings presents significant challenges under time and resource constraints, particularly when sound categories are imbalanced and only part of a dataset has been labeled. This paper describes the development and evaluation of a human--AI workflow used during annotation of the Kenya Forest Sounds Dataset (KFSD), using a project-stage working collection of 7,531 audio recordings across 21 categories. We investigated zero-shot classification, a feature-based ensemble, and a fine-tuned Audio Spectrogram Transformer (AST). The fine-tuned AST achieved 89.36\% accuracy, 85.98\% precision, 89.24\% recall, and 87.58\% F1 on held-out test data for binary classification of \emph{Person talking}. Because this result remained below the predefined 95\% threshold for autonomous annotation, model predictions were used as suggestions and subjected to human verification. Of the 5,026 initially unlabeled recordings, 2,337 were routed to a speech-verification queue and 2,689 remained for ordinary manual annotation. The verifier reviewed all 2,337 machine-labeled files and detected zero false positives, yielding 100\% verification accuracy. The observed average verification time was approximately 4.2 seconds per file. At an average rate of 35 seconds per file for manual annotation, the hybrid workflow reduced work on the remaining queue from approximately 48.9 to 28.9 person-hours, saving about 20.0 hours (41\%). The study documents the model-selection process, provenance-aware file organization, verification procedure, observed errors, practical limitations, and opportunities for extending AI assistance beyond one category.
\end{abstract}

\section{Introduction}
Environmental sound classification supports applications such as ecological monitoring, urban analysis, and security. Collecting recordings has become easier with low-cost recording devices, but producing high-quality annotations remains labor-intensive. Audio clips must be listened to, interpreted, and assigned to potentially overlapping or acoustically similar categories. This creates a substantial gap between the availability of recorded sound and the availability of labeled data suitable for machine learning \cite{piczak2015}.

This study concerns the Kenya Forest Sounds Dataset (KFSD), an environmental-audio collection associated with the Curbing Illegal Logging Patterns using Sound-Based Detection Techniques project. The project-stage working collection examined here comprises 7,531 recordings, approximately ten hours of environmental audio, and 21 sound categories. At an average rate of 35 seconds per file, manual annotation of the complete collection would require approximately 73 person-hours. Annotation was already underway when model development began: 2,505 files had human labels and 5,026 files were still unlabeled. This provided some training data but also created a need to prevent newly generated model predictions from overriding completed human work.

The core challenges were time constraints, quality consistency, class imbalance, concurrent work, and the need for validation given potential AI involvement. Some sound categories had as few as two samples, while others had more than 1,000. Traditional automated approaches were unsuitable because the 21 categories were domain-specific and not well represented in existing datasets, the collection was highly imbalanced, and some categories (for example, \emph{Person laughing}, with only two samples) had insufficient data for training. Field recordings also included background noise, variation in source distance, and occasional acoustic overlap. These conditions motivated a narrower, verifiable use of artificial intelligence.

The central research question is how much annotation work a binary audio model can assist with when its outputs are treated as suggestions requiring human confirmation, rather than as final ground truth. The study therefore examines not only classifier accuracy but also the design and efficiency of the human verification process.

The contributions are fourfold: (i) a documented comparison of alternative AI-assisted approaches for this imbalanced audio collection; (ii) a fine-tuned binary AST for detecting the \emph{Person talking} category; (iii) a provenance-aware workflow separating human labels, unreviewed model predictions, and files requiring fresh labeling; and (iv) a streamlined verification procedure together with its effect on annotation effort. The scope is intentionally limited to a single machine-assisted category; the remainder of the dataset continues through manual annotation.

\section{Related Work}
\subsection{Environmental audio classification and transfer learning}
Early approaches to environmental sound recognition relied on spectro-temporal descriptors and convolutional architectures \cite{piczak2015,hershey2017}. More recent pretrained audio models can transfer representations learned from large corpora to smaller target datasets. The Audio Spectrogram Transformer (AST), which processes audio spectrogram patches using a transformer architecture, has been evaluated on large-scale audio event classification tasks \cite{gong2021}. Its publicly available AudioSet-finetuned checkpoint offered a starting point for our experiments.

Contrastive Language--Audio Pretraining (CLAP) provides another direction by mapping sound and textual descriptions into a shared representation, enabling zero-shot inference through natural-language prompts \cite{wu2023}. Such methods are attractive when some target categories have insufficient examples for supervised learning. Transfer learning, more broadly, provides a basis for reusing models trained in one setting for another \cite{pratt1993}.

\subsection{Machine-assisted annotation and quality control}
Annotation systems can integrate predictions and human review rather than requiring a choice between fully manual and fully automated work. Klie et al. describe INCEpTION as a machine-assisted and knowledge-oriented interactive annotation platform \cite{klie2018}. Recent work on quality control in crowdsourced annotation highlights the importance of verification mechanisms, particularly when combining human and machine labels \cite{daniel2018}. Problems arising from inadequate attention to data work have also been documented in applied AI systems \cite{sambasivan2021}.

Our emphasis differs from building a new general-purpose recognition architecture. We document a constrained application of pretrained audio classification within an ongoing dataset-labeling project, including the operational decisions required when model accuracy is helpful but not sufficient for autonomous labeling. The distinction between classifier evaluation and human-reviewed deployment is important to the study.

\section{Kenya Forest Sounds Dataset (KFSD) and Annotation Setting}
\subsection{Audio collection and class distribution}
The project-stage KFSD working collection used in the annotation experiments contains 7,531 recordings sampled at 16\,kHz. Durations range from 0.11 to 5.02 seconds, with a mean duration of approximately 4.2 seconds. The annotation vocabulary includes 21 environmental sound categories. At the start of model development, 2,505 files (33.3\% of this working collection) had human labels, while 5,026 files (66.7\%) were awaiting annotation.

The project-stage working collection described in this study is distinct in scope from the publicly released KFSD record. The project repository documents the public release as containing 5,500 recordings \cite{kfsd2026,kfsdrepo2026}. Accordingly, 7,531 refers to the working collection used in the annotation experiments reported here, whereas 5,500 refers to the public dataset release documented by the project.

Class distribution in the initially labeled KFSD working subset was markedly uneven. \emph{Person talking} accounted for 1,056 labeled examples (42.2\% of the initially labeled subset), while \emph{Steel banging} and \emph{Person laughing} each had two examples (0.08\% of that subset). As a result, the amount of evidence available for individual classes varied substantially. This imbalance influenced both our choice of supervised target and our decision to avoid an immediate 21-class training strategy.

\begin{table}[t]
\centering\small
\caption{Overview of the KFSD project-stage working collection and annotation state at model development.}
\label{tab:dataset}
\begin{tabularx}{\linewidth}{@{}lX@{}}
\toprule
Characteristic & Value \\
\midrule
Working-collection recordings & 7,531 \\
Sound categories & 21 \\
Sampling rate & 16\,kHz \\
Recording duration & 0.11--5.02\,s (mean 4.2\,s) \\
Approximate recorded audio & 10 hours \\
Initially human labeled & 2,505 (33.3\%) \\
Initially unlabeled & 5,026 (66.7\%) \\
Frequent labeled category & Person talking: 1,056 examples \\
Rare labeled categories & Steel banging; Person laughing: 2 examples each \\
\bottomrule
\end{tabularx}
\end{table}

\subsection{Annotation constraints}
At an average rate of 35 seconds per file, manual annotation of the complete 7,531-recording collection would require approximately 73 person-hours. The annotation project also needed to accommodate work conducted in parallel: previously completed human annotations had to remain authoritative, machine-assisted cases required explicit review, and sounds rejected by the model still needed classification among all 21 target labels.

The public Kenya Forest Sounds Dataset (KFSD) release is available through the Zenodo record \cite{kfsd2026}. Dataset documentation is maintained in the KFSD repository \cite{kfsdrepo2026}. The accompanying annotation workflow, model configurations, and project code are available at \url{https://github.com/SCES-Makerspace-Lab/Auto-Sound-Annotation-2026}.

\section{Methodology}
\subsection{Selection of a model-assisted annotation task}
The original objective was to examine whether available pretrained audio classifiers could reduce the remaining manual annotation workload for the KFSD across its 21 categories. Four strategies were considered: CLAP-based zero-shot classification, an AudioSet-pretrained AST used without target-specific fine-tuning, an ensemble combining AST predictions with conventional audio features, and a fine-tuned AST. Based on the available labeled data and early experimental performance, the final supervised task was narrowed to binary detection of \emph{Person talking} versus \emph{Other}.

This category had more labeled examples than the remaining classes, providing a workable supervised training subset. The binary construction also avoided training independent classifiers for all 21 categories under a limited development schedule. The trade-off was intentional: predicted \emph{Other} files could not be assigned automatically to one of the other 20 labels and were therefore returned to ordinary manual annotation.

\subsection{Zero-shot CLAP experiment}
CLAP was initially selected because textual category descriptions could, in principle, represent classes for which little or no labeled training data was available \cite{wu2023}. We attempted to evaluate the 21-category vocabulary using natural-language prompts. In the Kaggle environment, however, installation produced dependency conflicts involving the versions of NumPy, SciPy, and Transformers required by the available components. The experiment was not completed. Rather than spending extensive time resolving environment issues, we pivoted to more stable alternatives and proceeded to an AST checkpoint with a working software configuration.

\subsection{AudioSet-pretrained AST baseline}
We loaded the checkpoint \texttt{MIT/ast-finetuned-audioset-10-10-0.4593}, which produces probabilities for 632 AudioSet classes. To approximate the target vocabulary, relevant AudioSet labels were associated with the 21 dataset categories. For instance, outputs corresponding to \emph{Speech}, \emph{Male speech}, and \emph{Female speech} were grouped under \emph{Person talking}. Probabilities associated with mapped AudioSet labels were aggregated to obtain dataset-level predictions.

On 300 labeled examples, this approach achieved 59\% accuracy. Inspection of predictions showed recurring confusions: birds were misclassified as speech, with high-frequency sounds confused; chainsaw sounds were misclassified as livestock, as both produce low-frequency rumbling; and wind was misclassified as silence or door sounds, as these are low-amplitude environmental sounds. The approach provided a quick baseline but did not offer adequate performance for automatic assignment of the specialized category labels. At a similar error rate on the 5,026 remaining files, approximately 2,060 incorrect labels would result.

\subsection{Feature-assisted zero-shot ensemble}
To investigate whether traditional signal descriptors could improve the pretrained model's predictions, we constructed a weighted ensemble comprising four components: AudioSet predictions (60\%), zero-crossing rate information (15\%), spectral centroid features (15\%), and mel-frequency cepstral coefficient variance (10\%). These features were included to provide complementary descriptions of signal characteristics not always captured by the direct mapping of pretrained AudioSet classes.

The ensemble achieved 68\% accuracy on the test set used for this approach. Although nine percentage points higher than the basic mapped-AST baseline, the result remained inadequate for autonomous annotation. An error rate of 32\% over the 5,026 remaining recordings would correspond to approximately 1,608 incorrect labels. We therefore did not use the ensemble to write final labels.

\subsection{Fine-tuned binary AST}
With 2,505 human-labeled KFSD examples available, we adapted the pretrained AST to the collection's acoustic conditions through supervised fine-tuning \cite{pratt1993}. We formulated a binary target: \emph{Person talking} was the positive class, and all other initially labeled categories formed the negative class. The data were divided into stratified training, validation, and test sets in a 70/15/15 proportion. Class-weighted cross-entropy was used to address imbalance.

For the KFSD binary classification task, the model was trained for 15 epochs with Adam at a learning rate of $3\times10^{-5}$ and a batch size of eight. Early stopping monitored validation F1 with a patience of ten epochs. The best reported validation accuracy was 92.82\% at epoch 11. On the held-out test data, the fine-tuned model achieved 89.36\% accuracy, 85.98\% precision, 89.24\% recall, and an F1 score of 87.58\%.

Before training, 95\% accuracy had been adopted as a threshold for considering autonomous labeling. Because the fine-tuned model remained below this threshold, its outputs were used only to propose likely speech files for subsequent human review. This preserved the benefits of model-assisted selection without treating unverified predictions as ground truth. Several factors limited performance: the \emph{Person talking} class includes whispers, shouts, conversations, and single words, which vary acoustically; recordings vary in source distance, background noise, and reverberation; the 1,056 speech examples still contained substantial acoustic diversity; and collapsing the other 20 categories into \emph{Other} removed important acoustic distinctions. We therefore treated machine predictions as initial suggestions, required systematic human verification, kept machine labels separate from human labels, and applied AI assistance only to \emph{Person talking}, not to all 21 categories. This was AI-assisted manual labeling for a subset of the data rather than fully automated labeling.

\subsection{Provenance-aware annotation workflow}
The KFSD annotation workflow maintained three separate file manifests to distinguish annotation states. The human-labeled manifest contained files with accepted human labels and took precedence if a file had also been processed by the model. The machine-labeled manifest contained previously unlabeled recordings predicted to be \emph{Person talking}; these entries represented pending review, not final accepted labels. The unlabeled manifest held recordings without a human annotation and without a positive speech prediction; these files required ordinary labeling among all candidate categories.

\begin{table}[H]
\centering\small
\caption{Purpose and priority of the three annotation manifests.}
\label{tab:manifests}
\begin{tabularx}{\linewidth}{@{}p{0.35\linewidth}X@{}}
\toprule
Manifest & Interpretation and next action \\
\midrule
\texttt{human\_labeled\_files.xlsx} & Human-annotated files; accepted human labels take precedence over model outputs. \\
\texttt{machine\_labeled\_files.xlsx} & Previously unlabeled files predicted as Person talking; each must be reviewed before acceptance. \\
\texttt{unlabeled\_files.xlsx} & Files with no human label and no positive model speech suggestion; assign one of the 21 classes manually. \\
\bottomrule
\end{tabularx}
\end{table}

For each recording, the assignment logic checks for an existing human label first. If one is present, the file enters the human-labeled manifest. If no human label exists but the model predicts \emph{Person talking}, the file enters the machine-verification queue. Otherwise, it remains in the unlabeled queue. The three-way partition prevents the same file from being assigned concurrently to multiple annotation queues and preserves label provenance. If validation failed, the entire machine-labeled batch would be rejected and the files moved to the unlabeled queue for ordinary manual annotation. Pseudocode for the routing rule appears in Appendix~\ref{app:algorithm}.

\begin{figure}[H]
\centering
\resizebox{0.96\linewidth}{!}{%
\begin{tikzpicture}[
  node distance=5mm and 9mm,
  box/.style={rectangle,rounded corners=2pt,draw=navy!65,fill=light,align=center,
      text width=32mm,minimum height=10mm,font=\footnotesize,inner sep=3pt},
  smallbox/.style={rectangle,rounded corners=2pt,draw=navy!50,fill=white,align=center,
      text width=33mm,minimum height=11mm,font=\footnotesize,inner sep=3pt},
  arr/.style={-{Stealth[length=2mm]},draw=navy!75,thin}]
  \node[box] (audio) {Incoming audio file};
  \node[box,right=of audio] (check) {Existing human label?};
  \node[smallbox,above right=4mm and 12mm of check] (human) {Human-labeled manifest};
  \node[box,below right=4mm and 12mm of check] (model) {Binary AST: speech?};
  \node[smallbox,right=12mm of model] (machine) {Machine prediction queue\\ Human verification};
  \node[smallbox,below=7mm of machine] (unlab) {Unlabeled queue\\ Manual 21-class annotation};
  \draw[arr] (audio) -- (check);
  \draw[arr] (check) -- node[above,sloped,font=\scriptsize]{Yes} (human);
  \draw[arr] (check) -- node[below,sloped,font=\scriptsize]{No} (model);
  \draw[arr] (model) -- node[above,font=\scriptsize]{Speech} (machine);
  \draw[arr] (model.south) |- node[pos=.32,below,font=\scriptsize]{Other} (unlab.west);
\end{tikzpicture}%
}
\caption{KFSD recording-routing logic. Human labels are retained; model-predicted speech enters a verification queue; all other unlabeled sounds follow conventional annotation.}
\label{fig:workflow}
\end{figure}
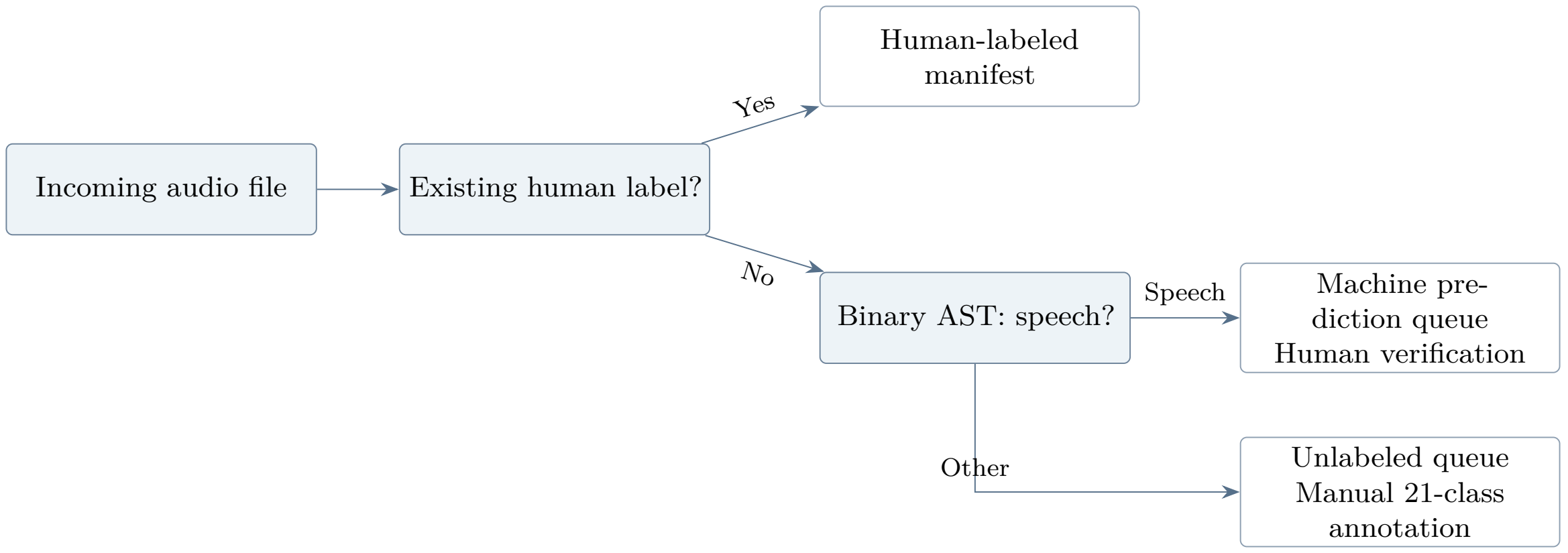

\subsection{Streamlined human verification}
Verification was designed around the fact that machine suggestions covered only a single target category. Instead of having the verifier repeatedly choose among 21 labels, each proposed speech file could be assessed with a binary question: does the recording actually contain \emph{Person talking}? This made it possible to move from full-category labeling to exception-focused checking for one subset of recordings.

A Python utility read the machine-prediction spreadsheet, located the corresponding sound recordings, and copied them into a designated review directory. The verifier then played all files in the verification folder sequentially using VLC (VLC media player). The files auto-advanced from one to the next, enabling continuous listening without manual file selection. During playback, the verifier noted only files that did not contain \emph{Person talking} (false positives), including empty or silent files and files containing other sound categories such as birds, wind, or vehicles. After playback, incorrectly predicted files were removed from the verification folder, annotated with their correct labels, and moved to the appropriate category.

Continuous playback was faster than opening files individually. The verifier only needed to track exceptions rather than every correct prediction, listening to all files ensured that false positives did not pass review, and the process could be applied to thousands of files with limited overhead. Every proposed speech file received auditory review. If validation failed, the entire machine-labeled batch would be rejected and the files moved to the unlabeled queue; human annotation could proceed independently without depending on AI predictions for dataset completion. The full file-organization script used for verification is reproduced in Appendix~\ref{app:script}.

\section{Experimental Setup and Evaluation}
\subsection{Computational environment}
Development and training were conducted in Kaggle Notebooks using an NVIDIA Tesla P100 GPU and 16GB RAM. The reported software environment consisted of PyTorch 2.0.1, Transformers 4.44.0, Librosa 0.11.0, and Python 3.8 for file organization. The AST checkpoint was \texttt{MIT/ast-finetuned-audioset-10-10-0.4593}. VLC (VLC media player) was used for sequential audio playback during manual verification.

\subsection{Classification metrics and evaluation protocol}
The experiments report classification accuracy, precision, recall, and F1 score. Accuracy is the proportion of correct classifications; precision is the fraction of positive predictions that are truly speech; recall is the fraction of speech examples recovered by the model; and F1 summarizes the balance between precision and recall. Precision was particularly important in this task because a false positive would place a non-speech recording into a speech-specific verification queue, although the subsequent review process remained responsible for final acceptance.

The mapped zero-shot AST was evaluated on 300 labeled examples. The fine-tuned binary AST used a stratified 70/15/15 split of the labeled set, with validation used during model development and the held-out portion used for test reporting. Production review was evaluated separately by listening to all recordings in the machine-positive verification queue. Classification performance and human-verification outcomes are therefore reported as distinct measurements.

\subsection{Annotation effort calculation}
Manual annotation was carried out at an average rate of 35 seconds per file. The observed average verification time was approximately 4.2 seconds per file: the verifier clicked the file and listened to the recording, with the time corresponding to its audio length, give or take 1--2 seconds. These times are used to compare the remaining 5,026-file queue with the hybrid procedure: 2,337 proposed speech files were reviewed, and 2,689 other files were annotated manually. The calculation concerns this remaining queue, not rework of the 2,505 files already labeled before the experiment.

The project progressed from four weeks of manual annotation to model development in week five, workflow validation in week six, and verification of machine predictions in week seven. The timeline and principal hyperparameters are also collected in Appendix~\ref{app:config}.

\section{Results}
\subsection{Comparison of classification approaches}
Table~\ref{tab:performance} compares the approaches with completed classification evaluations. The mapped zero-shot AST obtained 59.00\% accuracy. Incorporating conventional audio features improved accuracy to 68.00\%. Fine-tuning the AST on the labeled subset raised reported test accuracy to 89.36\%, an improvement of 30.36 percentage points relative to the mapped zero-shot baseline.

\begin{table}[t]
\centering\small
\caption{Reported classification performance for the evaluated approaches on the KFSD annotation task.}
\label{tab:performance}
\begin{tabular}{@{}lrrrr@{}}
\toprule
Method & Accuracy & Precision & Recall & F1 \\
\midrule
Zero-shot AST & 59.00\% & 59.04\% & 98.00\% & 73.69\% \\
AST + audio features & 68.00\% & 75.40\% & 95.00\% & 84.07\% \\
Fine-tuned AST & 89.36\% & 85.98\% & 89.24\% & 87.58\% \\
\bottomrule
\end{tabular}
\end{table}

The fine-tuned model achieved a peak validation accuracy of 92.82\% at epoch 11. Despite the improvement over zero-shot alternatives, it did not reach the 95\% threshold established for autonomous annotation. We therefore proceeded with selective model-assisted review rather than full automatic labeling. The CLAP option was not included in Table~\ref{tab:performance} because the dependency problem prevented a completed evaluation.

\subsection{Error patterns}
The zero-shot mapping showed errors associated with acoustic resemblance between classes, including birds categorized as speech, chainsaws as livestock, and wind as silence or door sounds. The fine-tuned binary classifier showed false positives for rhythmic footsteps, bird calls, and some breathy wind recordings. False negatives included distant speech, low signal-to-noise recordings, and cases in which speech overlapped other sounds.

The 15\% test split corresponds to 376 files, including 158 \emph{Person talking} clips. Based on this split and the reported test metrics, the corresponding confusion-matrix counts are inferred as 141 true positives, 23 false positives, 17 false negatives, and 195 true negatives. The false-positive patterns were person walking, where rhythmic footsteps were misclassified; birds, whose high-pitched chirping resembles the human voice; and wind, with its breathy noise patterns. The false-negative patterns were distant or quiet speech with a low signal-to-noise ratio, overlapping sounds in which speech occurred with background noise, and accented or non-standard speech. These patterns illustrate the challenge of recognizing speech across different distances, speaking styles, and background conditions.

Speech itself is acoustically diverse: the positive class may encompass conversation, shouting, whispering, or isolated words. Conversely, collapsing the other 20 sound classes into a single negative label combines acoustically heterogeneous recordings. Both factors constrain what the binary model can learn with the available sample count.

\subsection{Human review of model-positive recordings}
After applying the fine-tuned model to the 5,026 initially unlabeled KFSD working-collection recordings, 2,337 were routed to the \emph{Person talking} verification queue. The remaining 2,689 recordings proceeded to full manual annotation among the 21 categories. The verifier reviewed all 2,337 machine-positive files. All predictions were correct: zero false positives were detected, giving 100\% verification accuracy. This result exceeded our expectations and confirmed the model's reliability for this specific category in our acoustic environment.

This operational result is reported separately from model performance on the held-out test data, where classification was evaluated on both positive and negative examples. High-confidence predictions were included in \texttt{machine\_labeled\_files.xlsx}.

\subsection{Effect on annotation effort}
For the KFSD working collection, manual annotation at an average rate of 35 seconds per file would require 48.86 hours for the 5,026 remaining recordings (approximately 48.9 hours). In the hybrid workflow, verification of 2,337 proposed speech recordings at the observed average of 4.2 seconds per file required 2.73 hours. Manual annotation of the remaining 2,689 recordings required 26.14 hours. The hybrid total was 28.87 hours (approximately 28.9 hours), saving about 20.0 hours, or 41\%, relative to the manual baseline for the remaining queue.

\begin{table}[t]
\centering\small
\caption{KFSD annotation effort for the 5,026 files remaining after initial manual labeling.}
\label{tab:effort}
\begin{tabularx}{\linewidth}{@{}Xrrr@{}}
\toprule
Activity & Recordings & Sec/file & Hours \\
\midrule
Manual-only baseline & 5,026 & 35.00 & 48.86 \\
Hybrid: verify model positives & 2,337 & 4.20 & 2.73 \\
Hybrid: annotate remaining files & 2,689 & 35.00 & 26.14 \\
\midrule
Hybrid total & 5,026 & --- & 28.87 \\
Effort saved & --- & --- & 19.99 \\
\bottomrule
\end{tabularx}
\end{table}

The time saving applied to the \emph{Person talking} category; the other 20 categories remained subject to manual annotation. The savings were realized because all 2,337 machine predictions were verified correctly. Lower verification accuracy would reduce or eliminate the savings. Individual file review would require approximately 39.2 seconds per file (35 seconds for standard annotation plus 4.2 seconds of audio playback). Spreadsheet-based verification would require switching between the audio and spreadsheet, whereas the described approach used continuous playback and exception-only logging.

\section{Discussion}
\subsection{Why a limited AI role was selected}
Although the fine-tuned AST substantially outperformed both zero-shot approaches, we retained a conservative acceptance rule because the model did not meet its autonomy threshold. The chosen role of the model was to prioritize a recognizable class for efficient confirmation. This assignment acknowledges that a classifier can be useful in an annotation workflow without being reliable enough to create unreviewed labels.

The decision to select \emph{Person talking} reflected five considerations. First, this category had 1,056 labeled examples, while most other categories had fewer than 100. Second, human speech had a clearer acoustic signature than categories such as \emph{Wind} and \emph{Plastic wrapper}. Third, misclassifying sounds such as birds as wind was considered more problematic than a binary speech/non-speech error. Fourth, training 21 separate binary classifiers or one 21-class model would have required more time. Fifth, we chose a conservative approach: test AI assistance on one category before expanding. The AI assistance therefore covered one of the 21 categories and 2,337 of the 7,531 working-collection files; the other 20 categories continued through manual annotation. This was a pilot application, not full automation.

\subsection{Verification as part of the system}
The human review procedure was not merely a final accuracy check added to the model. It was a core component of the annotation system. File manifests protected existing human judgments, programmatic copying assembled the pending cases, continuous playback reduced per-file interaction, and exception logging limited data-entry burden. All machine predictions underwent systematic human review. In the reviewed batch, all 2,337 predictions were correct and zero false positives were detected, yielding 100\% verification accuracy. The design allowed model and manual work to proceed concurrently without replacing accepted human labels.

Machine labels were clearly marked and separated from human labels, the original audio remained accessible for re-annotation, and the decision rationale was documented. The verification process was streamlined but comprehensive. If validation failed, the entire machine-labeled batch would be rejected and the files moved to the unlabeled queue. Human annotation could then proceed independently, with no dependence on AI predictions for dataset completion.

\subsection{Differences between test performance and verified predictions}
The held-out binary classification assessment produced 89.36\% accuracy, while production verification of all 2,337 machine-positive predictions produced 100\% verification accuracy, with zero false positives. This discrepancy can be explained by four factors. First, the test set was deliberately stratified to include difficult edge cases. Second, production files came from similar recording conditions, providing greater acoustic-environment consistency. Third, high-confidence predictions were included in \texttt{machine\_labeled\_files.xlsx}. Fourth, files predicted as \emph{Person talking} likely had strong acoustic signals because of category prevalence.

The test metrics assess binary prediction on labeled held-out data, whereas the production result documents the human verifier's review of the selected machine-predicted speech files. The project experience shows that real-world accuracy may exceed test-set accuracy.

\subsection{Practical implications for annotation teams}
The project suggests that researchers with limited computing and annotation capacity can evaluate AI assistance incrementally. Defining an automation threshold at the outset made it possible to require human verification when model performance was below the desired level. Keeping a manual fallback allowed the project to continue when zero-shot approaches failed or when a proposed class could not be predicted reliably.

AI assistance may be useful when there are at least 1,000 labeled samples for the target category, the target category is acoustically distinct, an efficient verification workflow can be implemented, validation accuracy exceeds 85\%, and the project accepts that most of the work may still be manual. AI assistance should be avoided when labeled data are scarce (fewer than 500 samples), categories are highly similar (for example, \emph{Person talking} and \emph{Person singing}), verification tools or workflows are unavailable, dataset quality is paramount (as in medical or legal applications), or automation across many categories is required with limited training data.

Verification workflows can be optimized through continuous audio playback, logging only exceptions rather than confirming every prediction, programmatic file organization, and use of familiar tools such as VLC (VLC media player) or other audio players rather than custom interfaces. These practices reduce context switching and allow the verifier to focus on listening.

\section{Limitations}
The model-assisted procedure targeted only one of the 21 sound categories. Consequently, 2,689 of the initially unlabeled recordings still required ordinary multi-category labeling. For KFSD, the result is best interpreted as a pilot demonstration of selective assistance rather than an end-to-end automated annotation system.

Class imbalance and heterogeneous recording conditions limited model performance. Collapsing the 20 non-speech classes into \emph{Other} simplified training but discarded distinctions that might be relevant to general environmental sound recognition. The project did not investigate pitch shifting, time stretching, noise injection, or other augmentation techniques, and the model was not compared with alternatives such as pretrained audio neural networks (PANNs) \cite{kong2020} or wav2vec 2.0 \cite{baevski2020}. The AST was selected in part because a stable checkpoint and supporting documentation were available. Categories with fewer than 100 samples could not be reliably automated without techniques such as few-shot learning, which were not explored.

Operational limitations also remain. Concurrent manual and machine workflows require careful file synchronization and provenance management. Continuous playback and exception logging depend on audio-review functionality, so implementation may differ across annotation platforms. The observed average verification time was approximately 4.2 seconds per file (the audio length, with the verifier clicking the file and listening, give or take 1--2 seconds).

For the reviewed batch, the verifier found zero false positives among the 2,337 machine-positive recordings, yielding 100\% verification accuracy and exceeding expectations. This result is consistent with the project observation that real-world accuracy may exceed test-set accuracy. However, the observed effectiveness of the review queue in this recording environment should not be assumed to transfer unchanged to substantially different acoustic contexts. Additional categories, data sources, and larger evaluations would be required to establish broader applicability.

\section{Future Work}
Several directions follow directly from the pilot. First, separate models could be trained for categories with sufficient data (more than 500 samples), or hierarchical classification and multi-task learning could be investigated; this would extend assistance beyond speech while retaining explicit per-category evaluation. Second, active learning could prioritize cases where the model is uncertain so that additional human annotations improve model coverage efficiently \cite{settles2009}.

Third, few-shot methods, including prototypical networks \cite{snell2017}, may be useful for categories represented by only two to ten examples. Fourth, confidence calibration could be studied to determine whether high-confidence predictions are consistently correct under varying conditions; the original project proposed investigating confidence thresholds above 98\%. Fifth, dedicated verification interfaces could combine continuous playback, exception logging, and spreadsheet updates in a single tool. Finally, multicategory evaluation should examine how class-specific mistakes affect time savings and annotation quality as the scope of AI assistance grows.

\section{Conclusion}
We have described a human-verified annotation workflow developed for the Kenya Forest Sounds Dataset (KFSD), using a project-stage working collection of 7,531 environmental audio recordings in 21 sound categories. Model comparisons showed improvements from 59.00\% accuracy for the mapped zero-shot AST to 68.00\% for an audio-feature ensemble and 89.36\% for a fine-tuned binary AST. The fine-tuned model improved accuracy by 30.36 percentage points over the zero-shot AST but did not meet the study's 95\% threshold for autonomous annotation. It was therefore used to identify likely instances of \emph{Person talking} for mandatory human review rather than to assign final labels autonomously.

The workflow routed 2,337 recordings to a speech-verification queue and left 2,689 for ordinary manual annotation. The verifier found zero false positives among the 2,337 machine predictions, corresponding to 100\% verification accuracy; this result exceeded our expectations. At an average rate of 35 seconds per file for manual annotation and an observed average of 4.2 seconds per file for verification, the workload for the remaining queue decreased from approximately 48.9 to 28.9 person-hours, saving about 20.0 hours (41\%). The AI assistance applied to one of the 21 categories and affected 2,337 of the 7,531 working-collection files (approximately 31\% of the collection). The other 20 categories continued through manual annotation.

This work documents a pilot application rather than a fully automated solution. It shows that AI assistance can provide value even when imperfect, that verification workflows significantly affect realized time savings, that real-world accuracy may exceed test-set accuracy, and that modest improvements can still be worthwhile. The time savings depended on the verification outcome: lower verification accuracy would reduce or eliminate them. This work does not show how to fully automate annotation, how to achieve 95\% accuracy with limited data, how to scale AI to all categories in an imbalanced dataset, or how to obtain transformative cost reductions in annotation workflows. The documented experience may be particularly useful for academic research groups with limited budgets, pilot studies where dataset size is uncertain, projects with highly imbalanced category distributions, and research groups deciding whether AI assistance is worth the implementation effort.

The project began with the hope that AI could dramatically accelerate annotation. The reality was more nuanced: AI helped with one of the 21 categories, saved approximately 41\% of the effort for the remaining annotation queue, and required systematic verification of every prediction. This is not a failure but an honest outcome. Not every AI application achieves transformative results, and documenting limited successes is as valuable as reporting breakthroughs. The 100\% verification accuracy achieved suggests that well-tuned models on acoustically consistent data can be highly reliable, even when test-set accuracy suggests otherwise.

\section*{Data and Code Availability}
The Kenya Forest Sounds Dataset (KFSD) is publicly accessible through its Zenodo record \cite{kfsd2026}. Dataset documentation is available in the project repository \cite{kfsdrepo2026}. The annotation workflow, model configurations, and project code are available through the SCES Makerspace Lab repository: \url{https://github.com/SCES-Makerspace-Lab/Auto-Sound-Annotation-2026}.

\section*{Acknowledgments}
This study was supported by teams of data collectors: Ali Sheikh, Trevor Kingangi, and Monique Wamona; and data annotators: Lauryn Waruingi, Cicily Muendo, Arnold Ochieng', Shirleen Wanyingi, Lydiah Wangaru, Caldwell Kibe, and Michael Barasa. The larger research project, \emph{Curbing Illegal Logging Patterns using Sound-Based Detection Techniques}, was funded by Climate Change AI under grant number IG-2023-113 (\url{https://www.climatechange.ai/innovation_grants/2023/5}). We also acknowledge the supervisor who guided the hybrid workflow design and project scope, and the computing resources provided through Kaggle Notebooks.

\appendix
\section{Annotation Routing Pseudocode}\label{app:algorithm}
\begin{lstlisting}
For each audio recording:
    if a human label already exists:
        add the recording to human_labeled_files.xlsx
    else if the binary AST predicts Person talking:
        add the recording to machine_labeled_files.xlsx
        mark the prediction as requiring human verification
    else:
        add the recording to unlabeled_files.xlsx
        request normal annotation among all sound categories
\end{lstlisting}

\section{Audio File Organization Script}\label{app:script}
The Python script used in the verification procedure copies filenames listed in the model-prediction manifest into a dedicated folder. Paths in this reproduction are those documented in the original workflow and are adjustable for other environments.
\begin{lstlisting}[language=Python]
import pandas as pd
import shutil
from pathlib import Path

excel_file = "/mnt/c/Users/Lenovo/Downloads/machine_labeled_files.xlsx"
audio_folder = Path("/mnt/c/Users/Lenovo/Downloads/audio_files/standard")
target_folder = Path("/mnt/c/Users/Lenovo/Downloads/manual_review")
column_index = 0  # first Excel column

target_folder.mkdir(exist_ok=True)
df = pd.read_excel(excel_file)
missing = []

for filename in df.iloc[:, column_index].dropna():
    src = audio_folder / str(filename).strip()
    if src.exists():
        shutil.copy2(src, target_folder / src.name)
        print(f"Copied -> {src.name}")
    else:
        missing.append(filename)

print("\nDone")
if missing:
    print("\nMissing files:")
    for m in missing:
        print(m)
\end{lstlisting}

\section{Training Configuration and Project Timeline}\label{app:config}
\begin{table}[H]
\centering\small
\caption{Training parameters used for the fine-tuned binary AST.}
\begin{tabularx}{\linewidth}{@{}lX@{}}
\toprule
Parameter & Configuration \\
\midrule
Model & Audio Spectrogram Transformer (AST) \\
Task & Person talking vs. Other \\
Base checkpoint & \texttt{MIT/ast-finetuned-audioset-10-10-0.4593} \\
Optimizer & Adam \\
Learning rate & $3\times 10^{-5}$ \\
Batch size & 8 \\
Training duration & 15 epochs \\
Early stopping & Patience 10, monitoring validation F1 \\
Loss & Class-weighted cross-entropy \\
Data split & Stratified 70\% train / 15\% validation / 15\% test \\
\bottomrule
\end{tabularx}
\end{table}

The operational timeline was as follows: weeks 1--4 covered manual annotation of the initial 2,505 examples; week 5 covered model development and training; week 6 covered workflow validation and implementation; and week 7 covered review of the machine-predicted speech files.

\section{Verification Rules and Operational Lessons}
\small
The review protocol organized candidate recordings before playback, used continuous listening rather than repeated manual selection, and logged only files that did not contain the predicted category. The verifier listened to all machine-positive files and recorded only exceptions. In the reviewed batch, all 2,337 machine-positive files were correct: zero false positives were detected, yielding 100\% verification accuracy and exceeding expectations. Incorrectly predicted files could be removed, assigned their correct labels, and moved to the appropriate categories. If validation failed, the entire machine-labeled batch would be rejected and the files moved to the unlabeled queue; human annotation could proceed independently.

The operational lessons were to set a clear accuracy threshold before training; optimize the verification workflow because mandatory human review can be streamlined, with continuous playback reducing verification time from approximately 39.2 seconds per file to 4.2 seconds per file; recognize that binary classification is a shortcut rather than a solution for all 21 categories; document which files were labeled by humans and which were machine-labeled; recognize that real-world performance may differ from test-set performance and may exceed it; maintain a conservative scope because AI assistance to one category is a pilot rather than a comprehensive solution; and recognize that categories with fewer than 100 samples cannot be reliably automated without methods such as few-shot learning, which were not explored.

The project also identified two paths for improving verification itself: use a dedicated interface that synchronizes playback and exception recording; and introduce confidence-based sorting or calibration. In the reported implementation, the procedure remained deliberately simple, relying on tools already familiar to the verifier.

\end{document}